\documentclass[journal]{IEEEtran}

\usepackage{algorithm}
\usepackage{algorithmicx}
\usepackage[noend]{algpseudocode} 

\usepackage{cite}
\usepackage{acro}
\usepackage{graphicx}
\usepackage{float}
\usepackage{xcolor}
\usepackage{amsmath}
\usepackage{array}

\newcommand{\Fig}{Fig.}

\newcommand{\Tab}{Table}
\newcommand{\etal}{\textit{et al.}}

\DeclareAcronym{ROI}{
short=ROI,
long=region of interest,
}

\DeclareAcronym{IOU}{
short=IOU,
long=
,
}

\DeclareAcronym{cIOU}{
short=cIOU,
long=circle intersection over union,
}

\DeclareAcronym{DoF}{
short=DoF,
long=degrees of freedom,
}

\DeclareAcronym{CPL}{
short=CPL,
long=Center Point Localization,
}

\begin{document}
%
\title{Map3D: Registration Based Multi-Object Tracking on 3D Serial Whole Slide Images}
%
%
%

\title{Circle Representation for Medical Object Detection}
\author{Ethan~H.~Nguyen, Haichun~Yang, Ruining Deng,Yuzhe Lu, Zheyu Zhu, Joseph T. Roland, Le Lu, Bennett A. Landman, \IEEEmembership{Senior Member, IEEE}, Agnes B. Fogo, and Yuankai Huo, \IEEEmembership{Member, IEEE}

\thanks{This work was supported in part by NIH NIDDK DK56942(ABF) and NSF CAREER 1452485 (Landman) for support. \emph{(Corresponding author: Yuankai Huo. Email: yuankai.huo@vanderbilt.edu)}}

\thanks{E. H. Nguyen, R. Deng, Y. Lu, Z. Zhu, B. A. Landman, Y. Huo were with the Department of Electrical Engineering and Computer Science, Vanderbilt University, Nashville, TN 37235 USA.}

\thanks{H. Yang, J. T. Roland, A. B. Fogo were with the Department
of Pathology, Vanderbilt University Medical Center, Nashville,
TN, 37215, USA.}

\thanks{L. Lu was with PAII Inc., Bethesda MD 20817, USA.}

\thanks{\dag E.Nguyen and H.Yang contributed equally to the manuscript.}}

\maketitle

\markboth{Manuscript pre-print, October~2021}%
{Shell \MakeLowercase{\textit{et al.}}: Bare Demo of IEEEtran.cls for IEEE Journals}

\maketitle

\begin{abstract}
Box representation has been extensively used for object detection in computer vision. Such representation is efficacious but not necessarily optimized for biomedical objects (e.g., glomeruli), which play an essential role in renal pathology. In this paper, we propose a simple circle representation for medical object detection and introduce CircleNet, an anchor-free detection framework. Compared with the conventional bounding box representation, the proposed bounding circle representation innovates in three-fold: (1) it is optimized for ball-shaped biomedical objects; (2) The circle representation reduced the degree of freedom compared with box representation; (3) It is naturally more rotation invariant. When detecting glomeruli and nuclei on pathological images, the proposed circle representation achieved superior detection performance and be more rotation-invariant, compared with the bounding box. The code has been made publicly available: \color{red}{https://github.com/hrlblab/CircleNet}. 
\end{abstract}


\section{Introduction}
\IEEEPARstart{G}{lomerular} detection is widely used in renal pathology research for efficient and quantitative glomerular phenotyping~\cite{huo2021ai}. The glomerulus is the basic functional unit of the kidney to filter excess fluid and waste products from blood into urine. Therefore, precisely detecting and phenotyping glomeruli is critical for investigating various kidney diseases~\cite{d2013rise}. 

\begin{figure*}[t]
\begin{center}
\includegraphics[width=0.8\textwidth]{{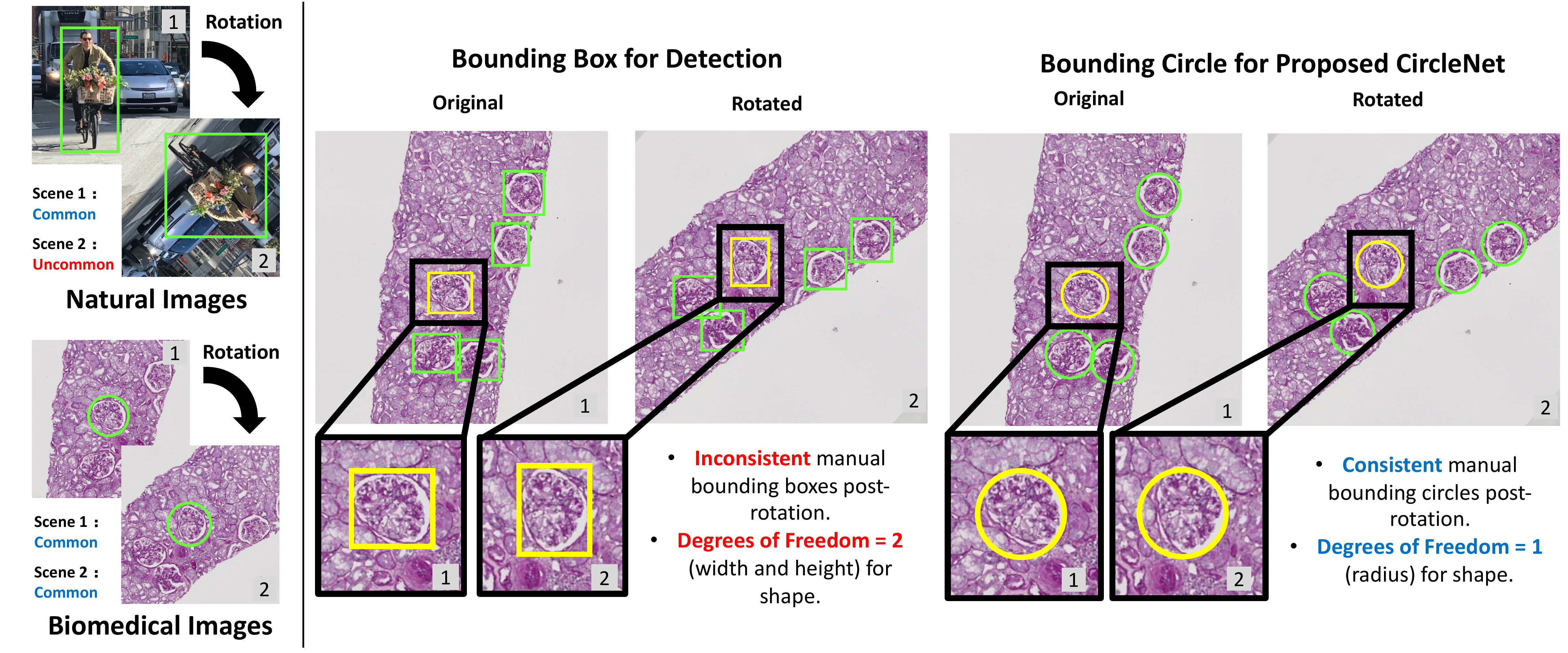}}
\end{center}
\caption{This figure showcases a comparison of the rectangular bounding box and CircleNet. The left panel shows how, unlike natural images, biomedical images can be commonly obtained with any angle of rotation. The right panel displays how the bounding box is not optimized for ball-shaped biomedical objects. The proposed CircleNet method produces a more consistent representation while requiring fewer degrees of freedom.} 
\label{Fig.1} 
\end{figure*}

While bounding box representation from the computer vision community is commonly utilized for detecting ball-shaped biomedical objects such as glomeruli~\cite{lo2018glomerulus, kawazoe2018faster, heckenauer2020real, rehem2021automatic}, such representation is not necessarily optimized (\Fig~\ref{Fig.1}). Certain biomedical images (e.g., microscopy imaging), unlike natural images, can be obtained and displayed at any angle of rotation of the same tissue. As a result, the traditional bounding box might yield inferior performance for representing such ball-shaped biomedical objects.

In this paper, we propose a simple circle representation for medical object detection and introduce CircleNet, an anchor-free detection framework based on the circle representation. After detecting the center location of the glomerulus, the proposed “bounding circle” requires one (radius) degree of freedom (DoF), while the bounding box needs two (height and width) DoF. The contributions of this study are in three key areas:

$\bullet$ \textbf{Circle Representation} We propose a simple circle representation for medical object detection that requires less DoF than the bounding box. We also introduce circle intersection over union (cIOU) as a metric for circle representation.

$\bullet$ \textbf{Optimized Medical Object Detection} To the best of our knowledge, CircleNet is the first anchor-free approach with optimized circle representation for detecting ball-shaped biomedical objects.  

$\bullet$ \textbf{Superior Detection and Rotation Consistency} Our proposed method, CircleNet, achieves superior detection performance and better rotation consistency compared to the bounding box. As demonstrated in \Fig \ref{Fig.1}, the tissue samples can be scanned with any arbitrary angles using WSI. Therefore, the better rotation consistency might lead to higher robustness for detecting the same objects from the same tissue, which would eventually improve the reproducibility of the image analytics. 

To evaluate the performance of the proposed CircleNet, three experiments were conducted. The first experiment measured its performance in detecting glomeruli on renal biopsies. The second experiment evaluated its performance in detecting nuclei on tissue samples. Lastly, the third experiment demonstrated the rotation consistency of the proposed CircleNet.

\textbf{Difference from Conference Version}: This work extends our previous conference paper~\cite{yang2020circlenet} with the following new efforts: A more comprehensive introduction, related works, and detailed data description are provided in this manuscript. The methodology is presented with more detailed mathematical derivations, experimental design, and hyper-parameter settings. To evaluate the generalizability of the proposed CircleNet beyond the glomerular detection, a new experiment on a different application (detection of nuclei) using a publicly available dataset~\cite{kumar2017dataset, kumar2019multi} is provided.

\section{Related Works}
\subsection{Object Detection}
Recent object detection methods based on convolutional neural networks (CNN) are divided into anchor-based and anchor-free object detectors. 
\subsubsection{Anchor-Based Methods}
Anchor-based object detection can be further categorized to two-stage~\cite{dai2016r, ren2015faster} and one-stage methods~\cite{liu2016ssd, lin2017focal}. 

Two-stage methods usually perform detection in two steps: $(i)$ region proposal and $(ii)$ object classification and bounding box regression. Faster-RCNN~\cite{ren2015faster} lays the groundwork for two-stage anchor-based detectors. Faster-RCNN~\cite{ren2015faster} consists of a region proposal network (RPN) and a prediction network (R-CNN)~\cite{girshick2014rich, girshick2015fast} that detects objects within each region. To tackle the challenge of proposing regions for objects of varying size and aspect ratios within the RPN, reference boxes called anchors were associated with a scale and aspect ratio. As the RPN checked each sliding window location, $k$ proposals were parameterized corresponding to the $k$ anchors. After Faster-RCNN, many algorithms were proposed to improve its performance, including different architectures~\cite{cai2016unified, dai2016r, cai2018cascade, lee2019me}, attention and context mechanism~\cite{bell2016inside, shrivastava2016contextual, liu2018structure, chen2018context}, different training strategy and loss function~\cite{najibi2016g, shrivastava2016training, wang2017fast, he2019bounding}, better proposal and balance~\cite{tan2019learning, pang2019libra}, feature fusion~\cite{lin2017feature}, and multi-scale training and testing~\cite{singh2018analysis, najibi2019autofocus}. While these two-stage detectors produce state-of-the-art results, they are often structurally complex and slower to inference.

One-stage methods eliminate the region proposal step and encapsulate all computations in a single network. With the introduction of SSD~\cite{liu2016ssd}, these types of methods have attracted academic attention for their high computational efficiency. SSD directly predicts object category and bounding box offsets by distributing the anchor boxes on multi-scale layers within a CNN. After SSD, various improvements have been suggested including redesigning the architecture~\cite{kim2018parallel, kong2018deep}, combining context from different layers~\cite{kong2017ron, fu2017dssd}, training from scratch~\cite{shen2017dsod, zhu2019scratchdet}, introducing different loss functions~\cite{chen2019towards, lin2017focal}, anchor matching and refinement~\cite{zhang2018single, zhang2020bridging}, and feature enrichment and alignment~\cite{liu2018receptive, zhang2018single, wang2019learning}. Currently, one-stage anchor-based object detectors obtain performance very close to two-stage anchor-based detectors but at faster inference speeds.

\subsubsection{Anchor-Free Methods}
Anchor-free methods remove the need for preset anchors. One approach to anchor-free detection is to localize several pre-defined or self-learned keypoints which generates the bounding boxes to detect objects (keypoint detection). CornerNet~\cite{law2018cornernet} detects a pair of keypoints using a single convolutional neural network: the top-left corner and bottom-right corner of a bounding box. By detecting the keypoints of the bounding box, the need for anchor boxes is eliminated. ExtremeNet~\cite{zhou2019bottom} improves on CornerNet by detecting five keypoints: the four corners of the bounding box and the center point of the object. However, CornerNet and ExtremeNet both require a computationally combinatorial grouping stage after keypoint detection which slows down each approach. So, CenterNet~\cite{zhou2019objects} approaches object detection by only extracting the center point of an object, removing the need for a grouping stage. 

\subsection{Medical Object Detection}
In the past, there have been many image processing methods proposed to detect biomedical objects. Historically, these methods strongly rely on human-designed imaging features such as edge detection~\cite{ma2009glomerulus,jung2010segmenting,esmaeilsabzali2012machine,filipczuk2013computer}, median filtering~\cite{kotyk2016measurement}, Histogram of Gradients (HOG)~\cite{kakimoto2014automated,kakimoto2015quantitative,kato2015segmental}, shape features~\cite{maree2016approach}, color-based and texture-based~\cite{ginley2017unsupervised}. However, these methods are limited by the lack of completeness of these hand-crafted features and their inability to generalize. 

In the last decade, deep convolutional neural network (CNN) based methods that rely on data-driven features have produced superior performance on detecting biomedical objects. Cireşan \textit{et~al.}~\cite{cirecsan2013mitosis} conducted mitosis detection in breast histology images by utilizing deep max-pooling convolutional neural networks. Temerinac-Ott \textit{et~al.}~\cite{temerinac2017detection} conducted glomerulus detection by integrating CNN performance on different stains. Gallego \textit{et~al.}~\cite{gallego2018glomerulus} proposed combining detection and classification, and other researchers~\cite{kannan2019segmentation,gadermayr2017cnn,ginley2019computational,bueno2020glomerulosclerosis,govind2018glomerular} integrated segmentation and classification. 

\subsubsection{Anchor-Based Methods}
Anchor-based methods, namely Faster-RCNN~\cite{ren2015faster}, have shown superior performance in computer vision tasks. Lo \textit{et~al.}\cite{lo2018glomerulus} and Kawazoe \textit{et~al.}\cite{kawazoe2018faster} applied the Faster-RCNN method to glomerulus detection which achieved state-of-the-art performance on the detection task. Mask-RCNN~\cite{he2017mask} was also adapted to detect the location of nuclei within a mask~\cite{johnson2018adapting}. However, anchor-based methods such as Faster-RCNN~\cite{ren2015faster} require anchors to be preset and refined throughout training, which typically yields lower flexibility and higher model complexity. Thus, detection methods without preset anchors resulting in simpler network design, fewer hyperparameters, and even superior performance have been the target of recent academic attention~\cite{law2018cornernet,zhou2019objects,zhou2019bottom}.

\subsubsection{Anchor-Free Methods}
Anchor-free methods such as CenterNet~\cite{zhou2019objects} have the potential to be faster and simpler. Feng \textit{et~al.}~\cite{feng2021automated} adapts ideas from  CenterNet~\cite{zhou2019objects} and RetinaNet~\cite{lin2017focal} to nuclei detection which achieves superior performance. However, such a method utilizes the traditional bounding box representation which is not necessarily optimized for circular biomedical objects.

\begin{figure}[t]
\begin{center}
\includegraphics[width=0.48\textwidth]{{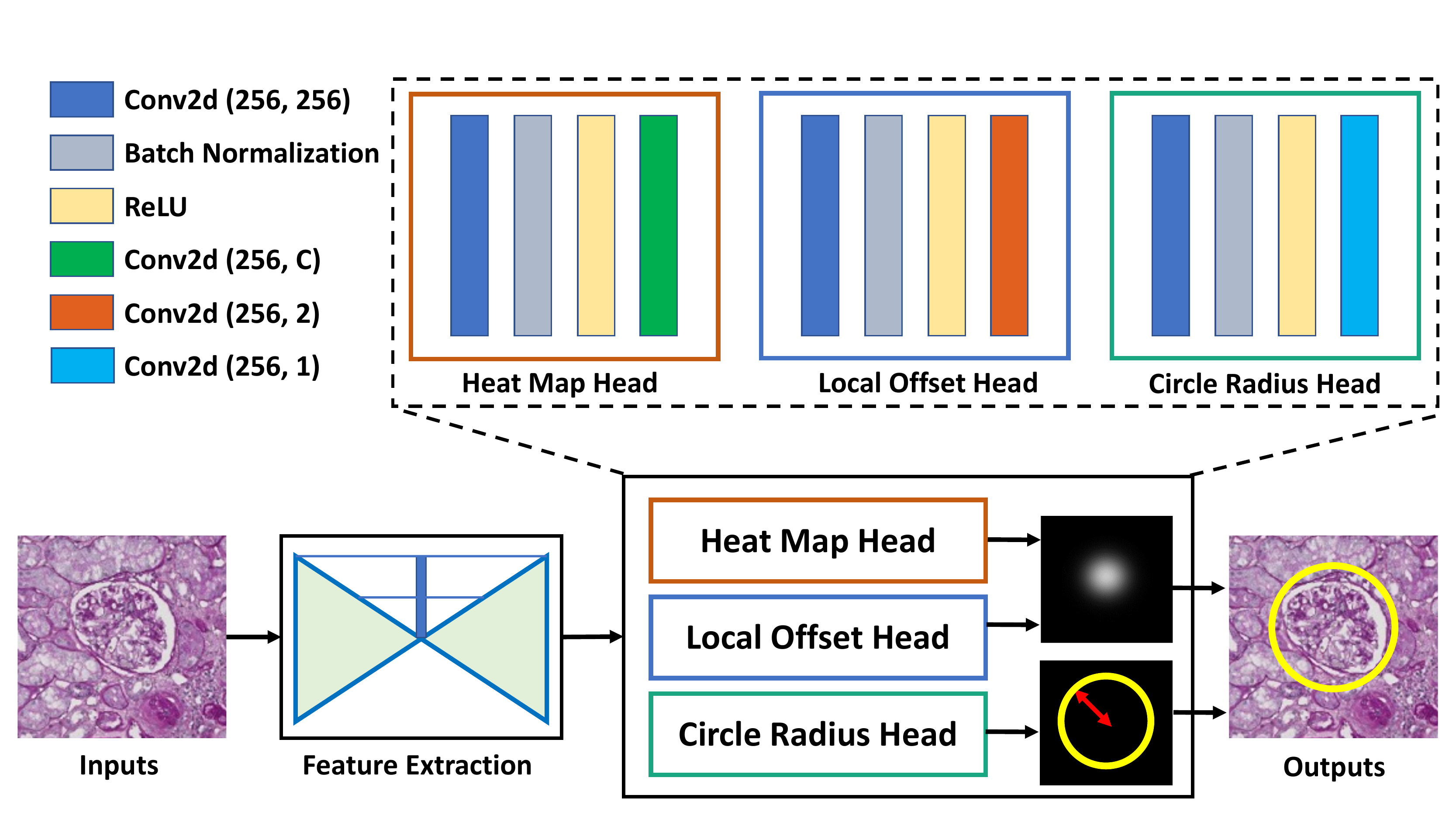}}
\end{center}
\caption{Overview of CircleNet. A backbone network serves as a feature extracter for the resulting three head networks. The heatmap and local offset head determines the center point of the circle while the circle radius head determines the radius of the circle.} 
\label{Fig.2} 
\end{figure}

\section{Methods}

\subsection{Anchor Free Backbone}
The overall framework of the proposed CircleNet is presented in \Fig ~\ref{Fig.2}. The network backbone is designed based on the anchor-free CenterNet implementation~\cite{zhou2019objects} for its high performance and simplicity. In addition, CenterNet is one of the most validated anchor-free methods. Many existing works are built upon CenterNet~\cite{peng2020deep, zhou2020tracking, li2020rtm3d}.

We follow Zhou \textit{et~al.} to define the key variables~\cite{zhou2019objects}. Let I be the input image where  $I \in R^{W \times H \times 3}$ with width $W$ and height $H$. From the network, the output is a heatmap $\hat Y \in [0,1]^{\frac{W}{R} \times \frac{H}{R} \times C}$ containing the center point localization of each object where C is the number of candidate classes and R is the downsampling factor of the prediction. The heatmap $\hat Y$ is expected to be 1 at the center of an object and 0 otherwise. Per convention~\cite{law2018cornernet, zhou2019objects}, ground truth of each object's center point is splat onto a heatmap $Y_{xyc} \in [0,1]^{\frac{W}{R} \times \frac{H}{R} \times C}$ using a 2D Gaussian kernel:

\begin{equation}
{Y_{xyc} = \exp\left(-\frac{(x-\tilde p_x)^2+(y-\tilde p_y)^2}{2\sigma_p^2}\right)}
\end{equation} 

where the $x$ and $y$ are the center point of the ground truth, $\tilde p_x$ and $\tilde p_y$ are the downsampled ground truth center point, and $\sigma_p$ is the kernel standard deviation. The training loss is $L_{k}$ penalty-reduced pixel-wise logistic regression with focal loss~\cite{lin2017focal}:

\begin{equation}
    L_k = \frac{-1}{N} \sum_{xyc}
    \begin{cases}
        (1 - \hat{Y}_{xyc})^{\alpha} 
        \log(\hat{Y}_{xyc}) & \!\text{if}\ Y_{xyc}=1\\
        \begin{array}{c}
        (1-Y_{xyc})^{\beta} 
        (\hat{Y}_{xyc})^{\alpha}\\
        \log(1-\hat{Y}_{xyc})
        \end{array}
        & \!\text{otherwise}
    \end{cases}
\end{equation}

where $\alpha$ and $\beta$ are hyper-parameters to the focal loss and $N$ is the number of keypoints~\cite{lin2017focal}. We empirically set $\alpha = 2$ and $\beta = 4$ experiments, following Law and Deng~\cite{law2018cornernet}.

\subsection{Center Point to Bounding Circle}
The top $n$ peaks are extracted from the heatmaps such that each peak's value is greater than or equal to its 8-connected neighbors. Let $\hat{P_c}$ be the set of $n$ detected center points where $\hat{\mathcal{P}} = \{(\hat x_i, \hat y_i)\}_{i = 1}^{n}$. The keypoint location of each object are given by integer coordinates $(x_i, y_i)$ from $\hat Y_{x_iy_ic}$ and $L_{k}$. Then, the offset $(\delta \hat x_i, \delta \hat y_i)$ is obtained from $L_{off}$. With center point $\hat{p}$ and radius $\hat{r}$, the bounding circle is defined as:
\begin{equation}
 \hat{p} = (\hat x_i + \delta \hat x_i ,\ \ \hat y_i + \delta \hat y_i). \quad \hat{r} = \hat R_{\hat x_i,\hat y_i}.
\end{equation}
where $\hat R  \in \mathcal{R}^{\frac{W}{R} \times \frac{H}{R} \times 1}$ is the prediction of the radius for each pixel location, optimized by 
\begin{equation}
    L_{radius} = \frac{1}{N}\sum_{k=1}^{N} \left|\hat R_{p_k} - r_k\right|.
\end{equation}
where $r_k$ is the ground truth of the radius for each circle object $k$. Finally, the overall objective is
\begin{equation}
    L_{det} = L_{k} + \lambda_{radius} L_{radius} + \lambda_{off}L_{off}.
\label{eq:total_loss}
\end{equation} Referring from~\cite{zhou2019objects}, we fix $\lambda_{radius} = 0.1$ and $ \lambda_{off} = 1$.

\begin{figure}
\begin{center}
\includegraphics[width=0.3\textwidth]{{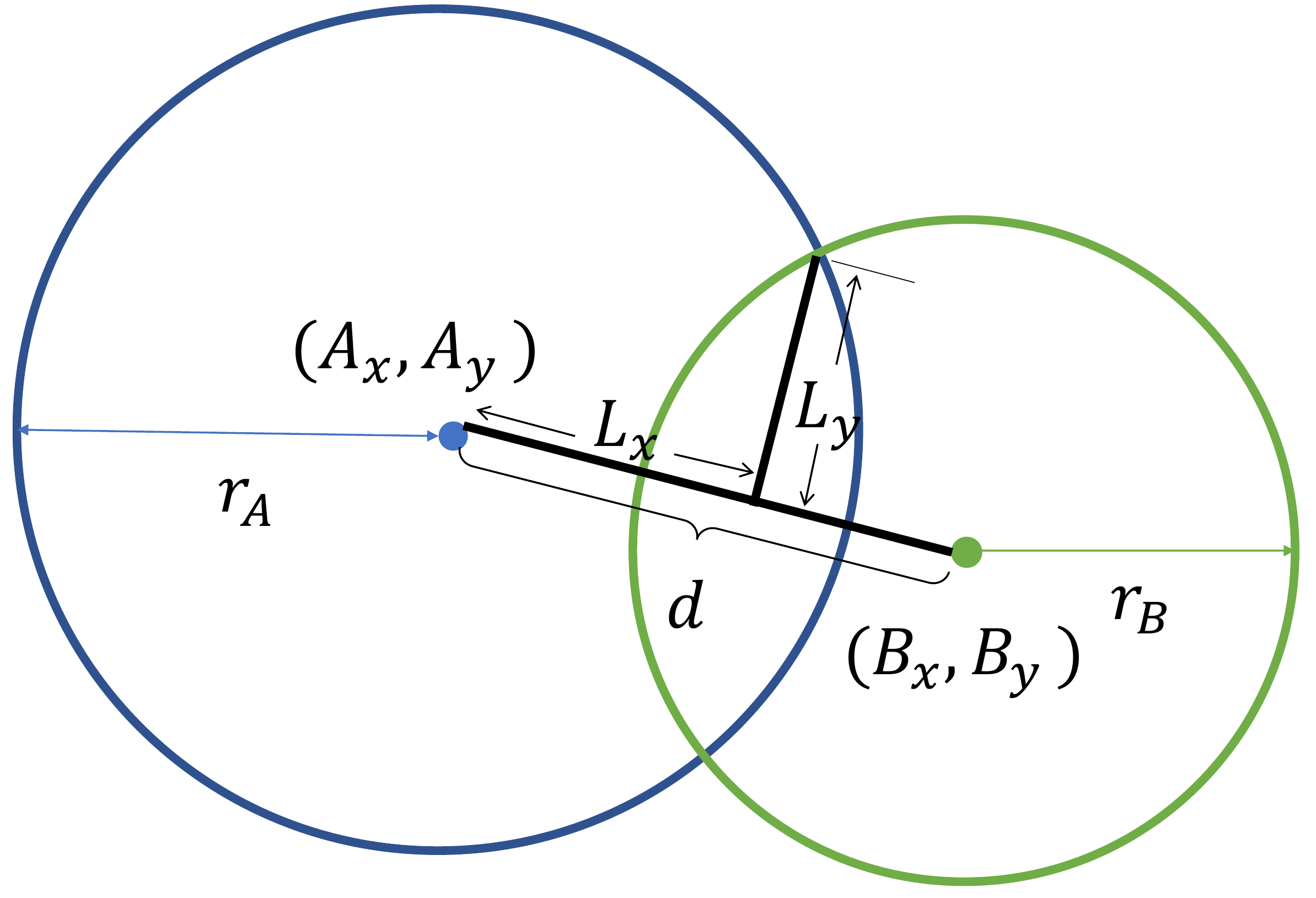}}
\end{center}
\caption{This figure showcases the parameters used to calculate circle IOU (cIOU).} 
\label{Fig.3} 
\end{figure}

\subsection{Circle IOU}
To measure the similarity between two bounding boxes, \ac{IOU} is the most popular evaluation metric in canonical object detection. The \ac{IOU} is defined as the ratio between the area of intersection and area of union. Analogously, to measure the similarity between two bounding circles, we introduce \ac{cIOU} as:
\begin{equation}
\textrm{cIOU} = \frac{\textrm{Area} \left( A \cap B \right)}{\textrm{Area} \left( A \cup B \right)}
\end{equation}
where $A$ and $B$ represent the two circles in \Fig~\ref{Fig.3}. The center coordinates of $A$ and $B$ are defined as $(A_x, A_y)$ and $(B_x, B_y)$ which are calculated as:
\begin{equation}
A_x = \hat x_i + \delta \hat x_i, A_y = \hat y_i + \delta \hat y_i
\end{equation}
\begin{equation}
B_x = \hat x_j + \delta \hat x_j, B_y = \hat y_j + \delta \hat y_j
\end{equation}
Then, the distance between the center coordinates $d$ is  defined as:
\begin{equation}
d = \sqrt{\left( B_x - A_x \right)^{2} + \left( B_y - A_y \right)^{2}}
\end{equation}
\begin{equation}
L_x = \frac{r_A^2 - r_B^2+d^2}{2d}, L_y = \sqrt{r_A^2 - L_x^2}
\end{equation}
Finally, the cIOU can be calculated from the following:

\begin{equation}
\begin{aligned}
\textrm{Area} & \left( A \cap B \right) = r_A^2\sin^{-1}\left( \frac{L_y}{r_A} \right) \\
& + r_B^2\sin^{-1}\left( \frac{L_y}{r_B} \right) - L_y\left(L_x + \sqrt{r_A^2 - r_B^2 +L_x^2}  \right) \   
\end{aligned}
\end{equation}

\begin{equation}
\textrm{Area} \left( A \cup B \right) = \pi r_A^2 + \pi r_B^2 - \textrm{Area} \left( A \cap B \right) 
\end{equation}

\section{Experimental Design}
\subsection{Data}
To obtain examples of glomeruli, whole slide images were captured from renal biopsies and annotated. The kidney tissue was routinely processed, paraffin-embedded, and $3\mu m$ thickness sections cut and stained with hematoxylin and eosin (HE), periodic acid–Schiff (PAS) or Jones. The samples were deindentified, and the studies were approved by the Institutional Review Board (IRB). 42 biopsy samples containing 704 glomeruli were used for training data, 7 biopsy samples containing 98 glomeruli for validation data, and 7 biopsy samples containing 147 glomeruli for testing data. Considering the ratio and size of a glomerulus with a patch~\cite{puelles2011glomerular}, the original high-resolution (0.25 $\mu m$ per pixel) whole scan images were downsampled to a lower resolution (4 $\mu m$ per pixel). Then, 10 random $512 \times 512$ patches per glomerulus image (original image contains at least one glomerulus as determined by the ground truth) were obtained as input images. Due to the foreground-background class imbalance, at least one positive object exists in all training patches following~\cite{lin2017focal}. Another rationale is from the widely used COCO dataset~\cite{lin2014microsoft}, in which only around 20 images do not have any objects given the total 328,000 images. Finally, these data formed a cohort containing 7040 training, 980 validation, and 1470 testing images.

\subsection{Experimental Design}
The implementation of CircleNet's detection and backbone networks followed the CenterNet's official PyTorch implementations. The COCO pre-trained model~\cite{lin2014microsoft} was used to initialize all models. All experiments were conducted on the same workstation with an 11 GB Nvidia 1080 Ti, Ubuntu 18.04, PyTorch 0.4.1, CUDA 9.0, and CUDNN 7.1. For data augmentation, random flip, cropping, and color jitter were used. The hyperparameters were 50 epochs, a learning rate of $2.5e-4$, and an Adam optimizer to adaptively alter the learning rate. Due to memory constraints, we set the batch size to 4.

As baseline methods, Faster-RCNN~\cite{ren2015faster}, CornerNet~\cite{law2018cornernet}, ExtremeNet~\cite{zhou2019bottom}, CenterNet~\cite{zhou2019objects} were chosen for their superior object detection performance. ResNet-50~\cite{he2016deep}, stacked Hourglass-104~\cite{newell2016stacked} network and deep layer aggregation (DLA) network~\cite{yu2018deep} were used as backbone networks for these different detection methods. For CircleNet, we followed the original implementation~\cite{yang2020circlenet} and use Hourglass-104 and DLA for the backbone networks. 

\subsection{Evaluation Metrics}
Mean average precision was the primary metric used to evaluate detection performance. For a given threshold IOU, average precision was obtained by calculating the area under the 101-point interpolated precision-recall curve. Then, the mean average precision ($AP$) is the mean of the average precision for IOU thresholds from 0.5 to 0.95 with a step size of 0.05. $AP_{50}$ is the average precision with an IOU threshold at 0.5. $AP_{75}$ is the average precision with an IOU threshold at 0.75. $AP_S$ is the mean average precision for small objects (area less than $32^2$). $AP_M$ is the mean average precision for medium objects (area between $32^2$ and $96^2$). Since no objects contained an area greater than $96^2$, the large mean average precision ($AP_L$) was not utilized.
\begin{table*}
\caption{CircleNet Glomeruli Detection Performance}
\centering
\begin{tabular}{ccccccc}
 \hline
Methods & Backbone & AP & $AP_{(50)}$ & $AP_{(75)}$ & $AP_{(S)}$ & $AP_{(M)}$\\
 \hline
Faster-RCNN\cite{ren2015faster}&ResNet-50 & 0.584 & 0.866 & 0.730 & 0.478 & 0.648\\
Faster-RCNN\cite{ren2015faster} & ResNet-101 & 0.568 & 0.867 & 0.694 & 0.460 & 0.633\\
CornerNet\cite{law2018cornernet}& Hourglass-104 & 0.595 & 0.818 & 0.732 & 0.524 & \textbf{0.695}\\
ExtremeNet\cite{zhou2019bottom} & Hourglass-104 & 0.597 & 0.864 & 0.749 & 0.493 & 0.658\\
CenterNet-HG\cite{zhou2019objects} &  Hourglass-104 & 0.574 & 0.853 & 0.708 & 0.442 & 0.649\\
CenterNet-DLA\cite{zhou2019objects} & DLA & 0.598 & 0.902 & 0.735 & 0.513 & 0.648\\
 \hline
CircleNet-HG (Ours) & Hourglass-104  & 0.615 & 0.853 & 0.750 & 0.586 & 0.656\\
CircleNet-DLA (Ours) & DLA & \textbf{0.647} & \textbf{0.907} & \textbf{0.787} & \textbf{0.597} & 0.685\\
 \hline
\end{tabular}
\label{table1}
\end{table*}

\begin{figure*}
\begin{center}
\includegraphics[width=0.9\textwidth]{{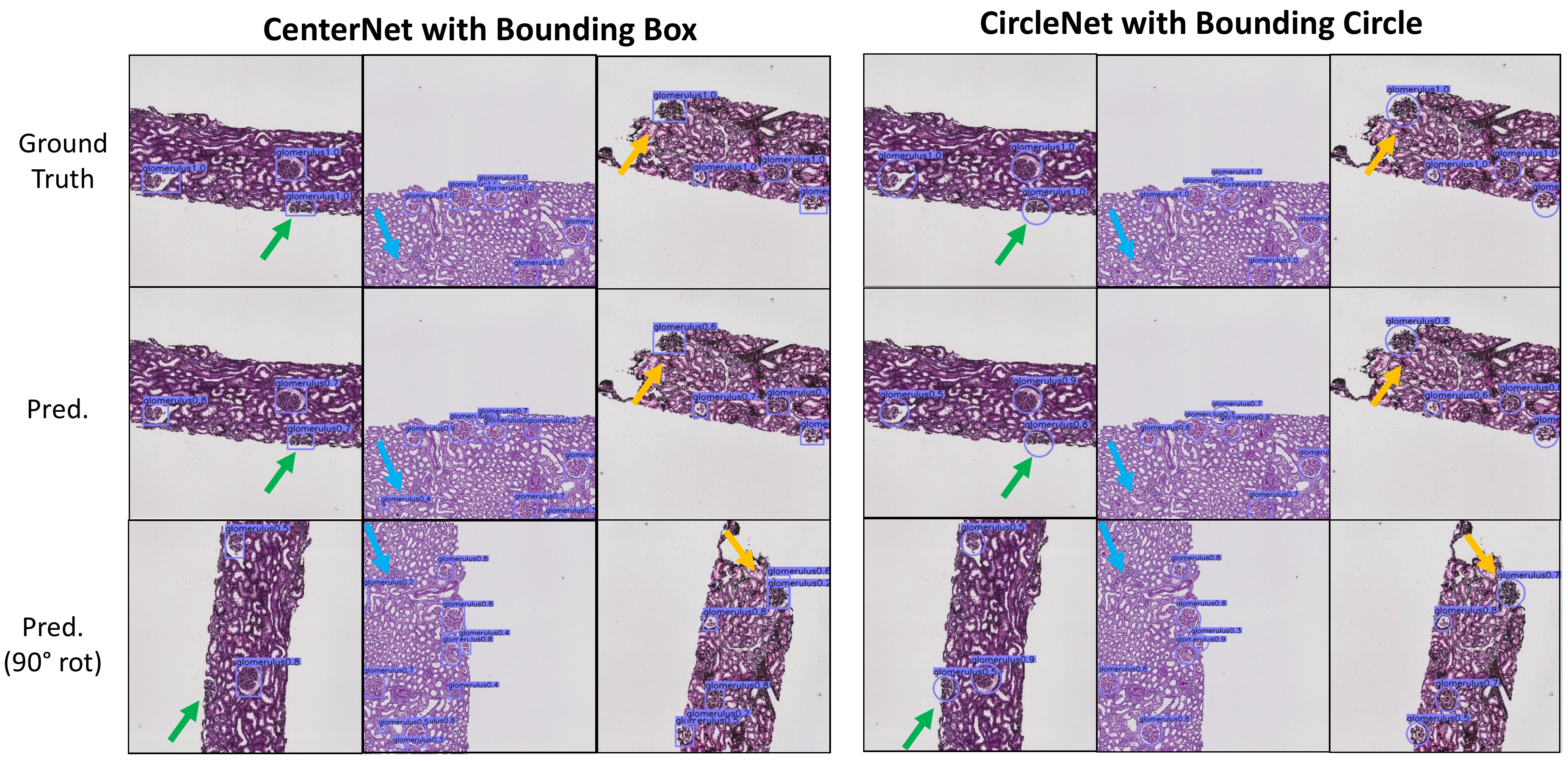}}
\end{center}
\caption{Qualitative comparison of glomerular detection results with confidence score $\ge$ 0.2. The confidence score was empirically selected for all experiments to balance the sensitivity and specificity.} 
\label{Fig.4} 
\end{figure*}

\section{Results}
\subsection{Glomerular Detection Performance}
As seen in {\Tab~\ref{table1}}, the proposed CircleNet method using the deep layer network (DLA) as a backbone outperforms the baseline methods on glomerular detection with a significant margin in all metrics except for $AP_M$. However, the proposed method still achieves the second-best performance for $AP_M$. In addition, when comparing CenterNet and CircleNet using the hourglass network (HG), the proposed CircleNet also performs better. Thus, across both backbone networks, CircleNet produces superior performance for glomerular detection.

A qualitative comparison between CenterNet and CircleNet can be seen in (\Fig~\ref{Fig.4}). As indicated by the arrows, CircleNet generally produces a representation more robust to rotation. 

\begin{figure}
\begin{center}
\includegraphics[width=0.48\textwidth]{{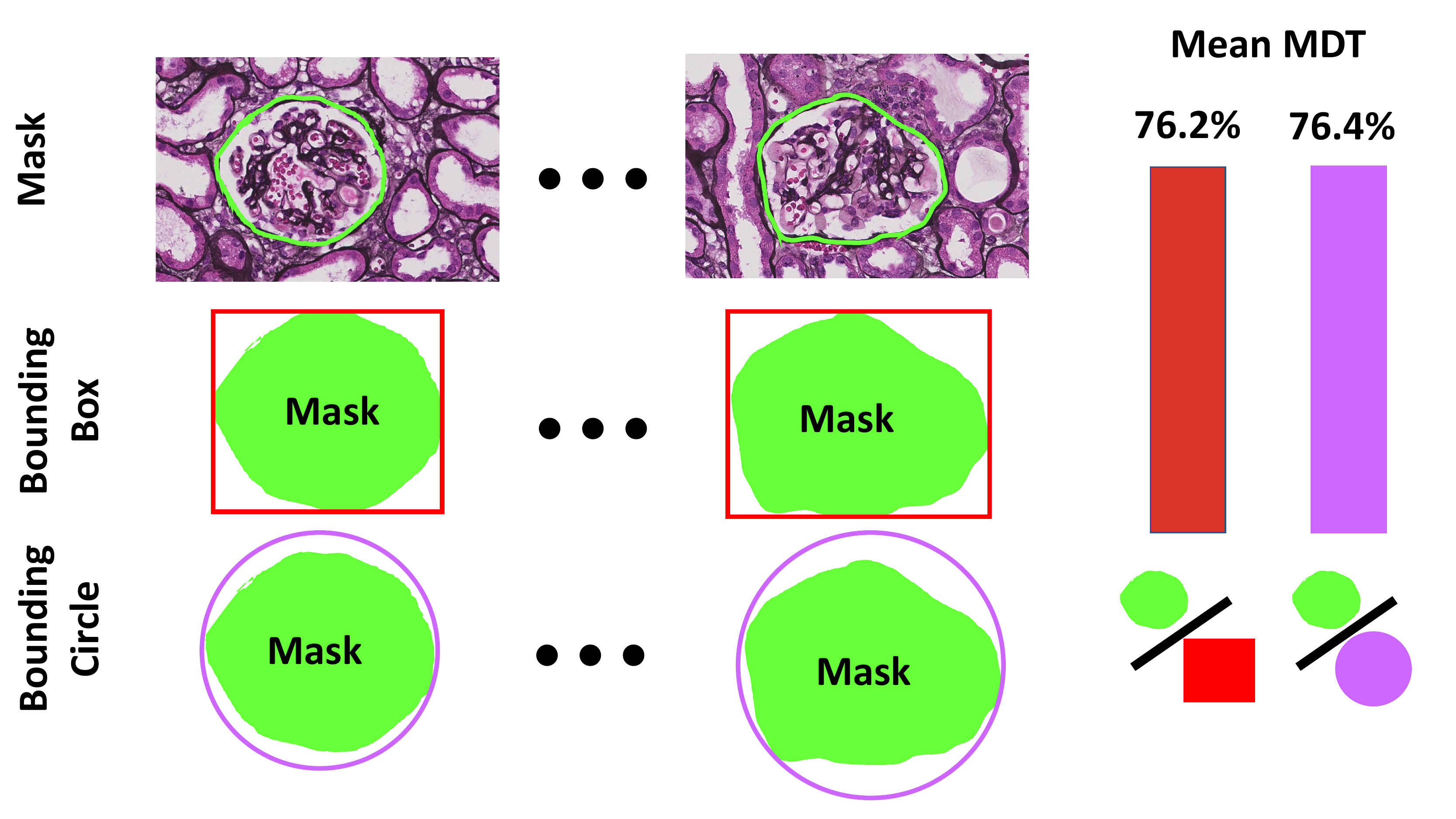}}
\end{center}
\caption{This figure showcases how the mean mask detection ratio (MDT) was calculated. The mask was originally traced on 50 randomly selected glomerulus from the testing dataset. The mean Mask Detection Ratio (MDT) was calculated from the average ratio between the mask area and bounding box/circle area. The mean MDT for both the rectangular box and circle representations were close.} 
\label{Fig.5} 
\end{figure}

\begin{figure}
\begin{center}
\includegraphics[width=0.48\textwidth]{{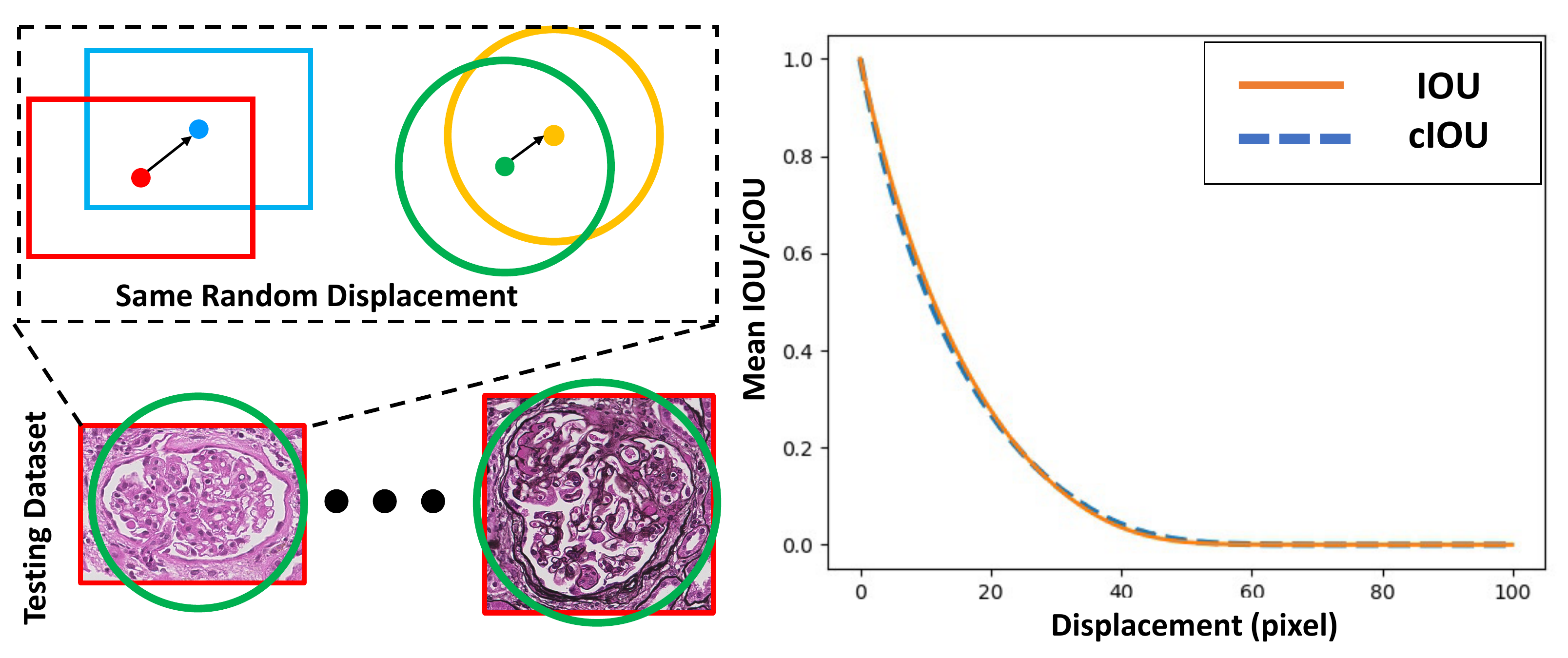}}
\end{center}
\caption{A comparison of IOU and cIOU. The bounding box/circle for every glomerulus in the testing dataset was shifted a random displacement (left panel). Then, the IOU and cIOU metrics were calculated to measure the similarity between the original and shifted bounding boxes/circles (right panel).} 
\label{Fig6}
\end{figure}

\subsection{Circle Representation and cIOU}
It was also investigated if the improved detection results of the bounding circle sacrificed its effectiveness for detection representation. To accomplish this, 50 glomeruli from the test dataset were manually annotated to obtain segmentation masks. Subsequently, the ratio between the mask area and bound box/circle area, called Mask Detection Ratio (MDT), were calculated for each glomerulus. As presented in the right panel of \Fig~\ref{Fig.5}, the box and circle representations both have comparable mean MDT, which demonstrates the bounding circle does not sacrifice effective detection representation and contributes to the improved detection performance.

Next, we compared the performance of the IOU and cIOU as metrics for similarity since metrics were used as overlap metrics for evaluating detection performance (e.g. $AP_{(50)}$ and $AP_{(75)}$). To measure their performance as similarity metrics, we simulated different detection results by adding random translations varying from 0 to 100 on the glomeruli in the testing dataset. To ensure a fair comparison, the same displacements were applied to each glomerulus (with bounding box and bounding circle representation). The results demonstrated in Fig. 6 show that cIOU behaves nearly the same as IOU, which validates cIOU as an overlap metric for the detection of glomeruli with the same random displacements. 

\begin{table}
\caption{Comparison of the Rotation Consistency between Bounding Box and Bounding Circle on Glomerular Detection}
\centering
\begin{tabular}{cccm{1.cm}<{\centering}}
\hline
Representation & Methods & Backbone & Rotation Consistency \\
\hline
Bounding Box & CenterNet-HG\cite{zhou2019objects} &  HG-104 & 0.833 \\
Bounding Box & CenterNet-DLA\cite{zhou2019objects} & DLA & 0.851 \\
\hline
Bounding Circle $\quad$ & CircleNet-HG (Ours) & HG-104  & 0.875 \\ 
Bounding Circle $\quad$ & CircleNet-DLA (Ours) & DLA & \textbf{0.886}  \\ 
\hline
\label{table2}
\end{tabular}
\end{table}

\subsection{Rotation Consistency}
An additional advantage of the bounding circle compared to the bounding box is better rotation consistency. We evaluated the consistency of the bounding box/circle by rotating the original test images by 90 degrees rather than an arbitrary angle to avoid the impact of intensity interpolation. Through this approach, the detected box/circle on rotated images were able to be converted to the original space. Furthermore, the rotation was only applied during testing and was not used as a data augmentation technique for training all methods. We calculated the rotation consistency by dividing the number of overlapped bounding boxes/circles (IOU or cIOU $>$ 0.5 before and after rotation) by the average number of total detected bounding boxes/circles (before and after rotation). This percentage of overlapped detection is named the "rotation consistency" ratio, where 1 means all boxes/circles overlapped while 0 means no boxes/circles overlapped. As seen in \Tab~\ref{table2}, the proposed CircleNet-DLA approach achieved better rotation consistency.

\begin{table*}
\caption{CircleNet MoNuSeg 2018 Detection Performance}
\centering
\begin{tabular}{ccccccc}
 \hline
Methods & Backbone & AP & $AP_{(50)}$ & $AP_{(75)}$ & $AP_{(S)}$ & $AP_{(M)}$\\
 \hline
Faster-RCNN\cite{ren2015faster} & ResNet-50 & 0.416 & 0.750 & 0.421 & 0.416 & 0.383\\
Faster-RCNN\cite{ren2015faster} & ResNet-101 & 0.409 & 0.775 & 0.372 & 0.410 & 0.339\\
CornerNet\cite{law2018cornernet}& HG-104 & 0.244 & 0.523 & 0.181 & 0.328 & 0.064\\
CenterNet-HG\cite{zhou2019objects} &  HG-104 & 0.447 & 0.846 & 0.427 & 0.451 &\textbf{0.395}\\
CenterNet-DLA\cite{zhou2019objects} & DLA & 0.399 & 0.826 & 0.315 & 0.403 & 0.338\\
 \hline
 CircleNet-HG (Ours) & Hourglass-104  & \textbf{0.487} & \textbf{0.856} & 0.509 & \textbf{0.499} & 0.337\\
CircleNet-DLA (Ours) & DLA & 0.486 & 0.855 & \textbf{0.516} & 0.499 & 0.305\\
 \hline
\end{tabular}
\label{table3}
\end{table*}

\begin{figure*}
\begin{center}
\includegraphics[width=0.9\textwidth]{{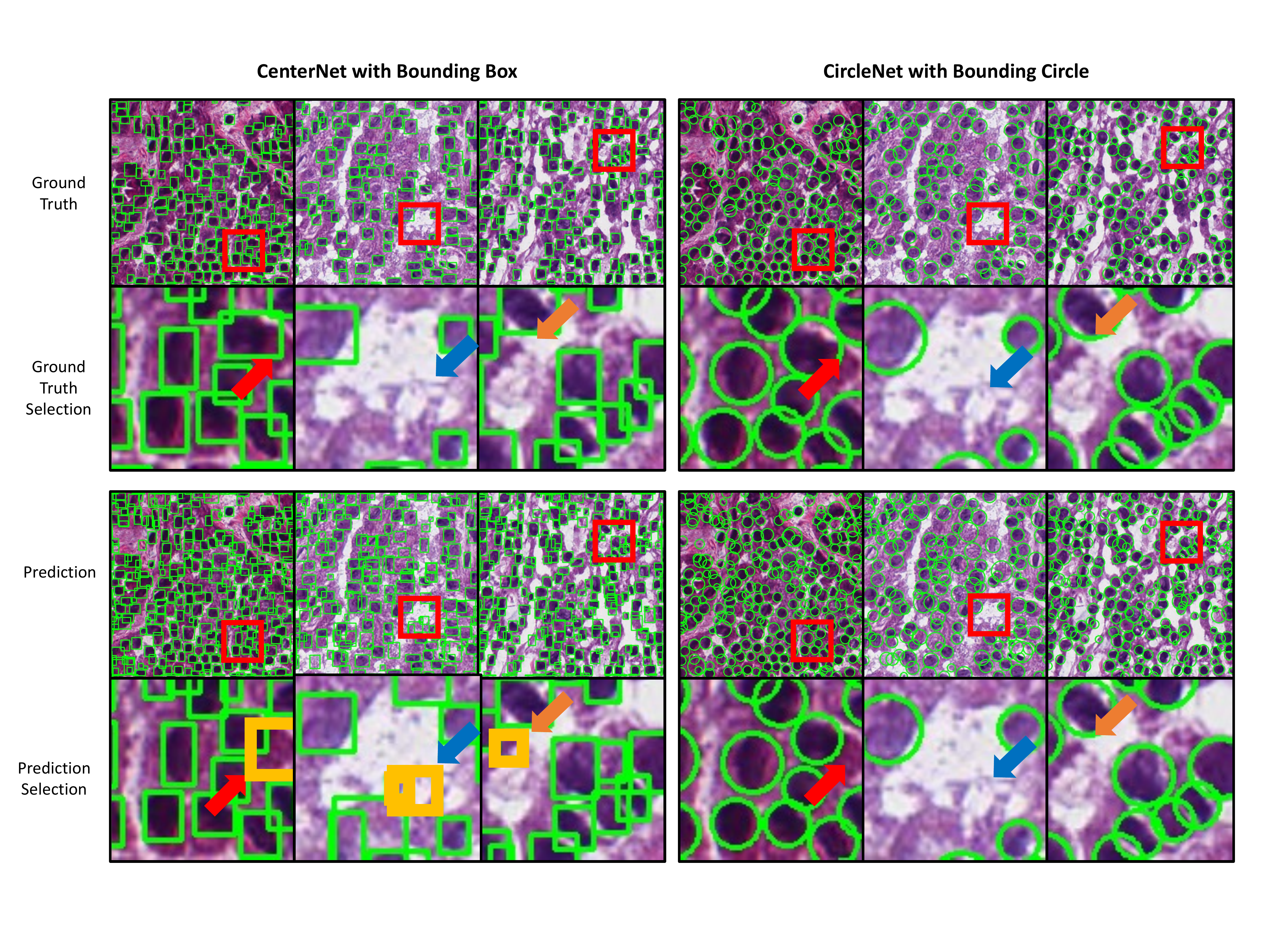}}
\end{center}
\caption{Qualitative comparison of nuclei detection results with confidence score $\ge$ 0.5. The confidence score was empirically selected for all experiments to balance the sensitivity and specificity. Within the orginal images, each red box indicates the location of each selection. Within each selection, a yellow box or circle indicates inconsistent detections. Arrows with the same color indicate the area of inconsistent detection across similar images.} 
\label{Fig.8} 
\end{figure*}

\begin{figure}
\begin{center}
\includegraphics[width=0.45\textwidth]{{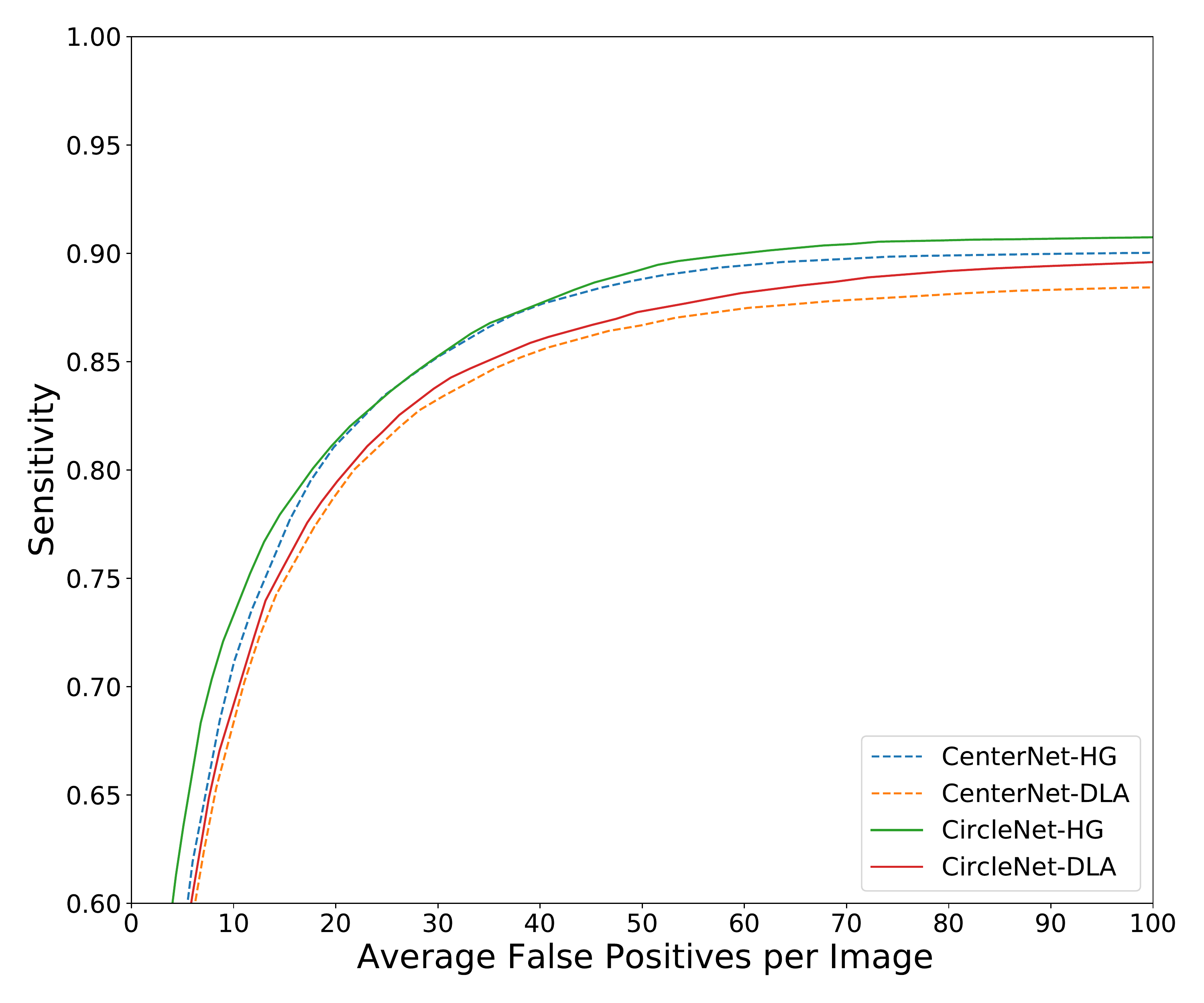}}
\end{center}
\caption{FROC curve of various methods on the test set of the nuclei dataset.} 
\label{Fig.9} 
\end{figure}

\begin{table}
\caption{Comparison of the Rotation Consistency between Bounding Box and Bounding Circle on MoNuSeg 2018}\label{tab4}
\centering
\begin{tabular}{cccm{1.cm}<{\centering}}
\hline
Representation & Methods & Backbone & Rotation Consistency \\
\hline
Bounding Box & CenterNet-HG\cite{zhou2019objects} &  HG-104 & 0.793 \\
Bounding Box & CenterNet-DLA\cite{zhou2019objects} & DLA & 0.853 \\
\hline
Bounding Circle $\quad$ & CircleNet-HG (Ours) & HG-104  & \textbf{0.891} \\ 
Bounding Circle $\quad$ & CircleNet-DLA (Ours) & DLA & 0.870  \\ 
\hline
\label{table4}
\end{tabular}
\end{table}

\subsection{Nuclei Detection Performance}
To validate our proposed method on another application, CircleNet was evaluated on a dataset from the 2018 Multi-Organ Nuclei Segmentation (MoNuSeg) Challenge~\cite{kumar2017dataset, kumar2019multi}. 

\subsubsection{Data}
The MoNuSeg challenge training/validation dataset includes 30 1000$\times$1000 tissue images containing 21,623 hand-annotated nuclear boundaries. Each image was sampled from a separate whole slide image of H\&E stained tissue at 40$\times$ magnification of several organs from \emph{The Cancer Genomic Atlas} (TCGA). A new testing dataset containing 14 1000$\times$1000 pixel images was also prepared using the same method as the training/validation data. This testing dataset contains lung and brain tissue images exclusive to the test dataset. 

From the 30 1000$\times$1000 pixel images in the training/validation dataset, 10 512$\times$ 512-pixel patches were randomly sampled from each image, generating 300 images for training/validation. Similarly, from the 14 1000$\times$1000 testing images, 140 images were obtained for testing. These data formed a cohort containing 200 training, 100 validation, and 140 testing images. 

While the original MoNuSeg training/validation dataset has relatively few images, those images contain more objects than the glomeruli dataset. Specifically, before data augmentation, the glomeruli dataset contains 802 glomeruli as compared to 21,623 nuclei in the MoNuSeg 2018 dataset. MoNuSeg 2018 is also publicly available.
\subsubsection{Approach}
The implementation and hyperparameters were the same as for glomerulus detection. 

\subsubsection{Results}
The results were evaluated using mean average precision similar to the metrics for glomerulus detection. As seen in \Tab~\ref{table3}, the proposed CircleNet-HG method outperforms the baseline methods on nuclei detection with a significant margin, except for $AP_M$. Additionally, when we compare CircleNet and CenterNet using the DLA network, CircleNet produces better results. A qualitative comparison between CenterNet and CircleNet can be seen in \Fig~\ref{Fig.8}. As displayed in \Tab~\ref{table4}, the proposed CircleNet-HG approach achieved better rotation consistency. Lastly, the FROC curve can be seen in \Fig~\ref{Fig.9}.
\section{Discussion}
In this study, we propose an anchor-free method, CircleNet, optimized for the detection of biomedical ball-shaped objects. Instead of using a bounding box representation, CircleNet uses a circle representation which is shown to offer superior detection performance and rotation consistency. The associated cIOU evaluation metric is shown to act similarly to IOU for bounding boxes. Thus, the results support that the circle representation indeed is more effective while requiring fewer degrees of freedom.

As seen in Table ~\ref{table2} and ~\ref{table4}, the proposed circle representation achieved better rotation consistency. One explanation for this result is that while length and width metrics are sensitive to rotation, radii are naturally more spatially invariant metrics.

Although only pathological image analysis is presented, we believe that circle representation is generalizable in radiology. Work by Luo \etal~\cite{luo2021scpmnet} has extended the circle representation into a sphere representation for lung nodule detection in 3D Computer Tomography scans. Compared with existing anchor-based and anchor-free methods, their anchor-free framework achieves superior performance. In addition, the sphere representation is verified to produce higher detection accuracy on long nodules than the traditional bounding box representation. Overall, their results support the potential generalizability of the circle representation within medical object detection.

One key limitation is that the circle representation may not be optimal for other types of shapes such as a stick-like shape or oval shape. Specifically, as seen in \Tab~\ref{tab4}, CircleNet performs worse than baseline methods for $AP_M$ which covers ~2\% of the objects in the testing dataset. Upon further analysis, larger nuclei tend to be more elongated and elliptical-like, unlike glomeruli which generally remain circular. An interesting potential improvement could be to add a second degree of freedom to the circle, transforming the circle into an ellipse. To define an ellipse representation, the network would predict the length of each axis within the ellipse in addition to the center point. In comparison to the circle representation, the ellipse representation may more effectively represent objects that are stick-like in shape.

While the circle representation achieves superior results within an anchor-free framework, a similar increase in performance may be obtained when adapting circle representation for anchor-based methods. The fact that the circle representation requires fewer degrees of freedom compared to the box representation could reduce the number of anchors are used. Instead of having bounding boxes of varying sizes and aspect ratios, an anchor-based approach with circle representation would only require bounding circles of varying radii. Further, having fewer anchors would reduce runtime and complexity. Overall, anchor-based methods may also benefit from using circle representations.

A promising application of CircleNet is within a detect-then-segment approach for the instance segmentation network of ball-shaped objects for Whole Slide Imaging (WSI). For instance, a single glomerulus from a 40$\times$ WSI can have a resolution of more than 1000$\times$1000 pixels. When using a standard segmentation approach like Mask-RCNN~\cite{he2017mask}, the corresponding feature maps are downsampled to 28$\times$28-pixel resolution which loses a substantial amount of information about the object. Therefore, applying CircleNet within a detect-then-segment approach may achieve superior segmentation performance for high-resolution WSI.
\section{Conclusion}
In this paper, we propose CircleNet, an anchor-free detection framework. The CircleNet method is optimized for ball-shaped biomedical objects, offering superior glomeruli and nuclei detection performance and rotation consistency. The circle representation and the cIOU evaluation metric were also comprehensively evaluated. The results show that, for detecting glomeruli and nuclei, the circle representation does not sacrifice effectiveness despite having fewer degrees of freedom compared with the traditional bounding box representation.

\section*{Acknowledgment}
E. H. Nguyen thanks Jonathan Ehrman for hardware support and Rose Rasty for editing and proofreading.

\bibliographystyle{IEEEtran}
\bibliography{main}

\end{document}